\documentclass[twocolumn]{article}

\usepackage{booktabs} % For formal tables

\usepackage{fancybox}
\usepackage{listings}

\usepackage{float}
\usepackage{footnote}
\usepackage{hyperref}
\usepackage{placeins}
\usepackage{booktabs} 	
\usepackage{tabularx}
\usepackage{array}
\usepackage{subfigure}
\usepackage{tikz}

\usepackage{subfigure}
\usepackage{multirow}
\usepackage{verbatim}
\usepackage{moreverb}

\usepackage{graphicx}
\usepackage{hyperref}
\usepackage{listings}
\usepackage{float}
\usepackage{graphics}
\usepackage{algorithm2e}
\usepackage{paralist}
\usepackage{amssymb}
\usepackage{amsmath}
\usepackage{listings}
\usepackage{setspace}
\usepackage[T1]{fontenc}
\usepackage{textcomp}
\usepackage[utf8]{inputenc}
\usepackage{csquotes}
\usepackage{newunicodechar} 
\usepackage{lstautogobble}
\usepackage{times}
\usepackage{latexsym}
\usepackage{lipsum}
\usepackage{listings}
\usepackage{authblk}

\usepackage{enumerate}

\newtheorem{definition}{Definition}

\newtheorem{example}{Example}

\usepackage{todonotes}

\definecolor{mygreen}{HTML}{00CC00}

   \newcommand\enrectangle[1]{%
  \tikz[baseline=(X.base)]
    \node (X) [draw, shape=rectangle, inner sep=0] {\strut #1};}

\newtheorem{req}{Requirement}

\begin{document}

\title{Principles for Developing a Knowledge Graph of Interlinked Events from News Headlines on Twitter }

%\author{Saeedeh Shekarpour}
%\affiliation{%
%  \institution{Knoesis Research Center}
%  \city{Dayton} 
%  \state{Ohio} 
%  \country{USA}
%}
%\email{saeedeh@knoesis.org}

\author[1]{Saeedeh Shekarpour\thanks{sshekarpour1@udayton.edu}}
\author[2]{Ankita Saxena}
\author[2]{Krishnaprasad Thirunarayan}
\author[2]{Valerie L. Shalin}
\author[2]{Amit Sheth}

\affil[1]{University of Dayton, Dayton, OH University}
\affil[2]{Knoesis Research Center,  Dayton, OH}

%\author{
%\authorblockN{Saeedeh Shekarpour\authorrefmark{2}, 
%Ankita Saxena\authorrefmark{2}, 
%Krishnaprasad Thirunarayan\authorrefmark{2}, 
%Valerie L. Shalin\authorrefmark{2}, 
%Amit Sheth\authorrefmark{2}}
%\IEEEauthorblockA{\\ \authorrefmark{2}Knoesis Research Center, USA, Email: \{FIRSTNAME\}@knoesis.org}
%}

%\IEEEauthorblockA{\\ \authorrefmark{1} University of Leipzig, Germany, 
%E-mail: \{alshargi,tsoru,quasthoff\}@informatik.uni-leipzig.de}

%
%\institute{Ohio Center of Excellence in Knowledge-enabled Computing (Kno.e.sis; http://knoesis.org)
%Wright State University, USA\\
%\email{\{saeedeh,ankita,tkprasad,valerie,amit\}@knoesis.org}}

\maketitle

%\IEEEcompsoctitleabstractindex

\begin{abstract}
The ever-growing datasets published on Linked Open Data mainly contain encyclopedic information. However, there is a lack of quality structured and semantically annotated datasets extracted from unstructured real-time sources.
In this paper, we present principles for developing a knowledge graph of interlinked events using the case study of news headlines published on Twitter which is a real-time and eventful source of fresh information.
We represent the essential pipeline containing the required tasks ranging from choosing background data model, event annotation (i.e., event recognition and classification), entity annotation and eventually interlinking events.
The state-of-the-art is limited to domain-specific scenarios for recognizing and classifying events, whereas this paper plays the role of a domain-agnostic road-map for developing a knowledge graph of interlinked events.
%inwe introduce a mechanism for representing n-ary relations and their arguments as a 
%background data model. 
%This representation leverages Levin's classification of English Verbs  in \cite{levin_english_1993} to support the use of unstructured text for constructing the background data model and capturing mentions of 
%n-ary relations.
%Then, we use learning approaches, employing proposed syntactic features derived from  parsing,
%to extract information respecting the data model.
%As a proof-of-concept, we follow a case study containing three distinct n-ary relations.
%The results of our experiments are promising and can be used to create timely and 
%structured news headlines dataset.
%\todoiteminlinedone{Amit}{Saeedeh}{Optional: somehow I see a little less of "why" in thisabstract. What does this do (for an application/use)?}

\end{abstract}

%\begin{IEEEkeywords}
%Event Knowledge Graph, Interlinking Events, Event Annotation, Event Data Model, News Headlines.
%\end{IEEEkeywords}

%############################################################
\section{Introduction}
%############################################################

Several successful efforts have led to publishing huge RDF (Resource Description Framework) datasets on Linked Open Data (LOD)\footnote{\url{http://lod-cloud.net/}} such as DBpedia \cite{dbpedia_iswc} and LinkedGeoData \cite{linkedgeodata}.  However, these sources are limited to either structured or semi-structured data. 
So far, a significant portion of the Web content consists of textual data from social network feeds, blogs, news, logs, etc.
Although the Natural Language Processing (NLP) community has developed approaches to extract essential information from plain text (e.g., \cite{StanforPOSTAgger,StanfordNE,StanfordCoreNLP}), there is convenient support for knowledge graph construction. 
Further, several lexical analysis based approaches extract only a limited form of metadata that is inadequate for supporting
applications such as question answering systems.  For example, the query
\textit{``Give me the list of reported events by BBC and CNN about the number of killed people in Yemen in the last four days''}, about a recent event (containing restrictions such as location and time) poses several challenges to the current state of Linked Data 
and relevant information extraction techniques. The query seeks ``fresh'' information 
(e.g., \textit{last four days}) whereas the current version of Linked Data is encyclopedic and historical,
and does not contain appropriate information present in a temporally annotated data stream. Further, the query specifies provenance 
(e.g., \textit{published by BBC and CNN}) that might not always be available on Linked Data. Crucially, the example query asks about a specific type of event (i.e., \textit{reports of war caused killing people}) with multiple arguments 
(e.g., in this case, location argument \textit{occurred in Yemen}). 
In spite of recent progress \cite{Ritter2012,RelatedEvents,j.websem466}, there is still no standardized mechanism for
(i)  selecting background data model, (ii) recognizing and classifying specific event types, 
(iii) identifying and labeling associated arguments (i.e., entities as well as relations), 
(iv) interlinking events, and (v) representing events.
%Although event extraction recently has been an open research area \cite{} in various communities such as data mining, NLP and Semantic Web, yet there is no clear suite representing the required steps and sketching upcoming challenges. 
In fact, most of the state-of-the-art solutions are ad hoc and limited. In this paper, we provide a systematic pipeline for developing  knowledge graph of interlinked events. As a proof-of-concept, we show a case study of headline news on Twitter.
The main contributions of this paper include:
%In fact, most of the state-of-the-art solutions are ad hoc and limited in a sense of generality. Since we observed an unclear systematic pipeline which can shed light on developing a knowledge graph of recent events which are interlinked with either historically related events or similar events published by other publishers, in this paper, we propose a suite which outlines the required steps for developing a knowledge graph of interlinked events from a real-time textual source using a case study of headline news on Twitter.
%First, we discuss the required background data model which is essential for representing events, their associated arguments and eventually interlinking events.
%The choice of a background data model can be either a reuse of the existing ontologies or developing an ontology from scratch.
%Thus, we review the existing popular data models and expose requirements for developing an ad hoc data model.
%As a proof-of-concept, we present a case study containing three distinct types of events in detail.
%We continue by sketching principles for annotating events as well as entities.
%Eventually, we discuss the necessity of interlinking events across time and media.

%#####################################
%########### figure
%#####################################
\begin{figure*}[htb]
\centering
\includegraphics[width=\textwidth]{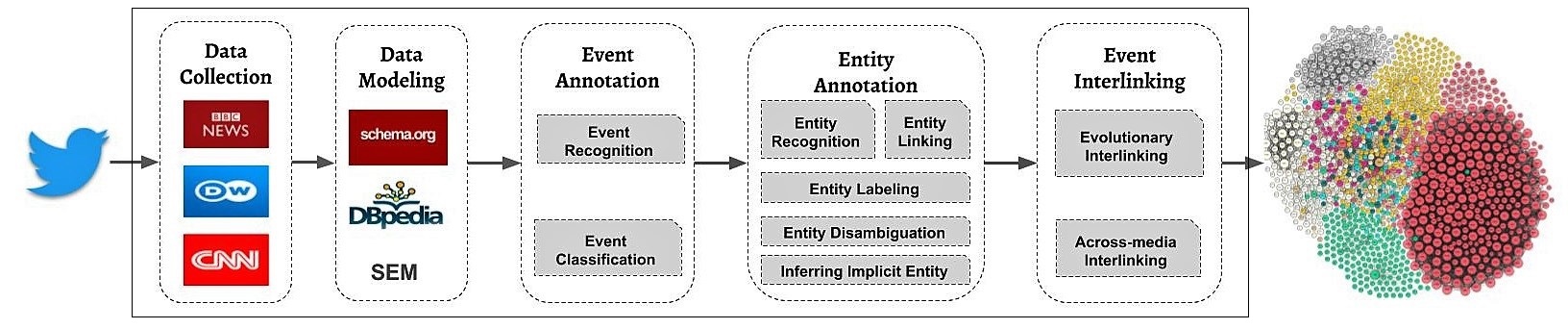} 
\caption{The pipeline of the required steps for developing a knowledge graph of interlinked events. }
\label{fig:Architecture}
\end{figure*}
%------------------------------------------

%to support the use of plain text for constructing the background data model and capturing mentions of n-ary relations.
%As a proof-of-concept, we present a case study containing three distinct n-ary relations.
%We represent the background data model for this case study in detail.
%Then, we propose the key features of a learning approach for recognizing and extracting associated entities (i.e., arguments) of n-ary relations. 
%These proposed features are mainly syntactic features derived from a dependency parse.
%Moreover, our  domain of interest is real-time tweets from multiple news agencies on Twitter. 
%News headlines, despite their brevity, are particularly informative and suitable 
%for gleaning events (formalized as n-ary relation instances). 
\begin{enumerate}[(I)]
\item The requirements for choosing a data model for representing and interlinking events.
\item Reviewing the-state-of-the-art data models for representing events. 
\item Incorporation of CEVO ontology for constructing the background data model and capturing fine-grained event types.
\item Presenting a comprehensive strategy for annotating events as well as entities.
\item Discussing the necessity of interlinking events across time and media.
\item Demonstration, using the news domain as a real-time data source.
(specifically, the stream of news headlines on Twitter).
\end{enumerate}
%To the best of our knowledge, this work provides the first framework for extracting triples based on a
%standarized representation of n-ary relations from a real-time data source (such as Twitter)\footnote{The demo and datasets will be published on the project homepage at \url{http://wiki.knoesis.org/index.php/HeadEx}.}.
The remainder of this paper is organized as follows.
Section \ref{sec:problemstatement} is dedicated to notation and problem statement.
Section \ref{sec:outline} outlines the required steps for developing a knowledge graph of interlinked events. 
Section \ref{sec:relatedwork} frames our contribution in the context of related work.
Section \ref{conclusion} concludes the paper with suggestions for future work.

%############################################################
\section{Notation and Problem Statement}
\label{sec:problemstatement}
%############################################################

A tweet of a news headline contains a sequence of words $t_{i}=(w_1,...,w_m)$.
\autoref{tab:tweetsamples} provides samples of news headlines on Twitter with provenance information such as publisher and publishing date.  These were sampled for the type of embedded event discussed below.
We aim to create an RDF knowledge base for such news headlines. 
An RDF knowledge base $K$ consists of a set of triples 
$(s, p, o) \in R \times P \times (R \cup L)$, where $R = C \cup P \cup I$ is the union of all 
RDF resources ($C, P, I$ are respectively a set of classes, properties and instances), and $L$ is a set of literals ($L \cap R = \emptyset$).
We aim to extract rich set of triples $(s_i,p_i,o_i)$ from each tweet $t_{i}$  in the stream of news headline tweets (as discussed below), and populate an \emph{event knowledge graph} $K_{event}$. Formally, the extraction task can be captured as $T \longrightarrow K_{event}$ where $T=\{t_{1},t_{2},...,t_{l} \}$ is the stream of news headline tweets and 
$K_{event}$ is a knowledge graph of events (where a tweet $t_i$ is mapped to a single event).
We address three main challenges on the way: (1) agreeing upon a background data model (either by developing or reusing one), (2) annotating events, associated entities as well as relations, (3) interlinking events across time and media, and (4) publishing triples on the event knowledge graph according to the principles of Linked Open Data. 
%In this paper, we shortly present the whole required pipeline and specifically elaborate on the available data models for publishing an event knowledge graph. 

\begin{table}[hptb]
	\centering
\begin{tiny}
\begin{tabular}{ l|l|l}
\hline
\textbf{Agency} 				&  \textbf{Date} 			& \textbf{News Headlines Tweets} 	   \\ \hline

\multirow{3}{*} {CNN }	 			&	16/3/16	&  	no1.  Michelle Obama tells \#SXSW crowd: I will not run for president  \\
     								&   26/2/16	&	no2. Instagram CEO meets with @Pontifex to discuss "the power of images to unite people" 	 	 	 \\  	
								&   14/3/16	&	 no3. Chemical accident in Bangkok bank kills eight people 	 		 \\ 	 \hline

\multirow{3}{*} {BBC }	   	&	14/3/16  		&  no4. State elections were "difficult day," German Chancellor Angela Merkel says  	  \\ 
 						&   10/3/16  			 & no5. Pope Francis visits Cuba and Mexico	 \\ 	
						    &   24/2/16 		 & no6. Storms kill at least three in Virginia	  \\	\hline 
						 		
\multirow{3}{*} {NYT}	 		&   10/3/16    &    no7. Obama and Justin Trudeau announce efforts to fight climate change		\\	
							&   10/3/16 	&   no8. Pope to meet leader of Russian Orthodox Church for first time in nearly  \\ 	
							&   10/3/16 	&	no9. 2 air force pilots from United Arab Emirates  killed when warplane crashed over Yemen\\  \hline

\end{tabular}
\end{tiny}
\caption{ Samples of news headlines from different publishers on Twitter.}
\label{tab:tweetsamples}
\end{table}

\section{Outline of The Required Steps}
\label{sec:outline}
Here, we outline the required steps for developing a knowledge graph of interlinked events.
Figure \ref{fig:Architecture} illustrates the high-level overview of the full pipeline.
This pipeline contains the following main steps, to be discussed in detail later.
(1) Collecting tweets from the stream of several news channels such as BBC and CNN on Twitter.
(2) Agreeing upon background data model.
(3) Event annotation potentially contains two subtasks (i) event recognition and (ii) event classification.
(4) Entity/relation annotation possibly comprises a series of tasks as (i) entity recognition, (ii) entity linking, (iii) entity disambiguation, (iv) semantic role labeling of entities and (v) inferring implicit entities.
(5) Interlinking events across time and media.
(6) Publishing event knowledge graph based on the best practices of Linked Open Data.

%our solution to address challenges (1) and (2) discussed above.
%This section is divided into two parts. The first part is dedicated to necessary requirements and foundations for representing background data model.
%The second part proposes key features for recognizing and extracting relations and their associated entities from tweets.

\subsection{Background Data Model}
\label{subsect:DataModel}
An initial key question is \emph{``What is the suitable background data model (serving as the pivot) for extracting triples associated to an event?''} Contemporary approaches to  extracting RDF triples capture entities and relations in terms of  binary relations \cite{NEL2015,BOA2013,banko2008tradeoffs}. 
We divide the current triple-based extraction approaches into two categories: (i) those that (e.g., \cite{NEL2015}) follow the pattern $(?e_1,p,?e_2)$ to leverage existing relations (i.e., properties) $p$ in the knowledge base to find the entities $e_1$ and $e_2$ for which the relation $(e_1,p,e_2)$ holds.
For example, for the relation \texttt{plays} holds between an athlete and his/her favorite sport, and NELL\footnote{\url{http://rtw.ml.cmu.edu/rtw/}} extracts the triple \texttt{seve ballesteros plays golf} for two entities \texttt{seve ballesteros} and \texttt{golf}, and
(ii) others that (e.g., \cite{rdflive,BOA2013}) utilize the pattern $(e_1,?p,e_2)$ to leverage the entities available in the knowledge graph (i.e., $e_1,e_2$) to infer new relations (e.g., $p$) that either did not exist in the knowledge base or  did not hold between the entities $e_1,e_2$.  
For example, \cite{rdflive} initially recognizes named entities in a given sentence and then, by inferring over domains and ranges of properties in DBpedia,  assigns an appropriate property between the recognized entities. Given an entity (e.g. \texttt{Garry Marshall}) with type \texttt{director} associated with a known movie (e.g. \texttt{Pretty woman}), it infers the property \texttt{dbpedia:director}\footnote{prefix as dbpedia:\url{http://dbpedia.org/property/}} from background ontology between the two recognized entities \texttt{Garry Marshall} and \texttt{Pretty woman}.
So far, supervised and unsupervised learning approaches have been applied for these extractions, which rely
on the use of a large number of specific lexical, syntactical and semantic features.
We assume that each news headline maps to an event modeled by an n-ary relation that can be captured by generating multiple triples.
An $n$-ary relation is a relation with n arguments $R(e_1,e_2,...,e_n)$.  For example,  a binary relation triple 
$(e_1,p,e_2)$ can be rewritten as $p=R(e_1,e_2)$.
%This study is the first step in our research agenda; so here we do not learn patterns. 
%Instead we predefine a couple of patterns with various number of arguments and try to extract triples using those patterns. 
Thus, the first challenge concerns the suitable background data model for representing various types of events and their associated entities by simulating n-ary relationships in terms of binary relationships.

Considering our case study, news headlines are often one single sentence (potentially accompanied by subordinate clauses) along with a link directing to the body of the news report\footnote{Although twitter content is often noisy and contains informal language, news headlines published by well-known news agencies are formal and well-written.}.  
In spite of its brevity, headline tweets provide dense and significant 
information.  Various entities appear in the embedded core message (the latter commonly as verb phrase), including aspects that indicate temporal properties, location and agent.
For example, consider the tweet no.2 in \autoref{tab:tweetsamples} that will serve as a running example: \textit{Instagram CEO meets with @Pontifex to discuss "the power of images to unite people"} that contains several entities related to the verb phrase \emph{`meet'} and are distinguished by separating boxes as \enrectangle{Instagram CEO} \enrectangle{meets with} \enrectangle{@Pontifex} \enrectangle{to discuss "the power of images to unite people"}. 
The general intuition is that a core verb (i.e., relation) heads each headline tweet accompanied by
multiple arguments (i.e., entities).
The number of entities $n$ depends on the type of relation but location and time are generic default arguments for any relation $n>2$. 
Thus, the core chunk (verb phrase) corresponds to the meet event and the remaining chunks of the given tweet likely function as  dependent entities of this event. 
For instance, in the running example, the chunk \enrectangle{meets} corresponds to the event $event_{meet}$ with the following recognized entities as associated arguments:
%\[event_{meet}(-time, -location,\texttt{Instagram CEO}, \texttt{@Pontifex},\texttt{to discuss...})\] 

\begin{equation}
event_{meet} =\begin{cases}
\texttt{generic type} = \texttt{event} \\
\texttt{specific type} = \texttt{meet} \\
\texttt{time} = - \\
\texttt{location} = - \\
\texttt{entity 1} = \texttt{Instagram CEO} \\
\texttt{entity 2} =\texttt{@Pontifex} \\
\texttt{entity 3} =\texttt{discussing "the power} \\ \texttt{of images to unite people"}
\end{cases}
\label{eq:wf}
\end{equation}

In this example, the temporal, as well as location arguments of $event_{meet}$, are absent. 
Consistent with linguistic theory, not all arguments are always present for each occurrence of an event.

%\todoiteminlinedone{Saedeg}{TKP}{Saeede: I added two more arguments, YOU SHOULD EITHER REQUIRE AT LEAST TWO ARGUMENTS TO CAPTURE SPATIO-TEMPORAL CONTEXT, OR 
%JUST MAKE A PASSING REMARK THAT NORMALLY PEOPLE ASSOCIATE AT LEAST TWO ARGUMENTS BESIDES THOSE 
%REQUIRED TO CAPTURE PROVINCE OR MEANING OF THE RELATIONSHIP BEING CAPTURED.}
%\todoiteminlinedone{VLS}{Saeedeh}{Saeede:thanks, makes sense.....Careful here---this is complex technical territory.  I wrote something that is a reasonable approximation}

%We limit our case study to three classes of English verbs. These three classes represent three distinct n-ary relations differing in terms of meaning as well as associated entities.
%The next step is representing these three relations in our background data model.   

The RDF\footnote{\url{https://www.w3.org/RDF/}} and 
OWL (Web Ontology language)\footnote{\url{https://www.w3.org/TR/owl-features/}} 
primarily allow binary relations, defined as a link between either two entities or an entity and its associated property value.
However, in the domain of news, we often encounter events that involve more than two entities, and hence require n-ary relations. % (ii) attributes inherently accompanying an relation.
The W3C Working group Note\footnote{\url{https://www.w3.org/TR/swbp-n-aryRelations/}} suggests two patterns for dealing with n-ary relations. 
We prefer the first pattern that creates $n+1$ classes and $n$ new properties to represent an n-ary relation.
% which is called \emph{reifiedrelations}\footnote{\url{https://www.w3.org/wiki/PropertyReificationVocabulary}}. 
We formally define a \emph{generic event class} representing all categories of events (n-ary relations) and then, use 
a template-based definition for any subclass of the generic event.
This enables the representation of specific types of events (e.g. meet event). 
% by representing associated entities as well as properties.

\begin{definition} [Class of Generic Event]
A generic event class refers to any event that can involve {\em n}  multiple entities.
In other words, the Generic Event Class denoted by $GEvent$ abstracts a relation among n entities.  
\end{definition}

\begin{definition} [Class of `X' Event]
`X' Event denoted by $C_X$  is a subclass (i.e. specific type) of the class $GEvent$, i.e., $C_X \sqsubset GEvent$. Conceptually it refers to events sharing common behavior, semantics, and consequences.
 \end{definition}

In the following, we provide requirements on the data model for developing a knowledge graph of interlinked events.

\begin{req}[Inclusion of Generic Event]
An event data model minimally includes the definition of the generic event while including the specific event as optional.
\end{req}

\begin{req}[Inclusion of Provenance]
\label{req:prov}
The provenance of each event must be represented within the data model. 
\end{req}

\begin{req}[Inclusion of Entity Type]
\label{req:type}
The type of each entity associated with a given event must be represented within the data model. 
This type can be fine-grained or coarse-grained.
\end{req}

\begin{req}[Inclusion of Properties]
\label{req:relation}
For any given entity $e_i$ associated with a given event $r_i \in C_X$, a property (i.e., binary relation) $p$ between the entity $e_i$ and the event $r_i$ must be represented within the data model. 
Thus, for the given pair $(e_i,r_i)$, either the triple $(e_i,p,r_i)$ or the triple $(e_i,p,r_i)$ is entailed in the RDF graph of $K_{event}$.
\end{req}

%\begin{req}[Relation Entailment]
%For any given event instance $r_i \in C_X$ and any of its associated entities $e_i$, the tripple 
%Every instance of this class $r \in C_i$ is 
%The Semantic Connected Component (SCC) of an entity $e$ in an RDF graph $G$ under a consequence relation $\models$ is defined as $SCC_{G,\models}(e) = \{ (e,p,o) \mid G \models \{ (e,p,o) \} \} \cup \{ (e,p,o) \in G\} $.
%If the graph and consequence relation is clear from the context, we use the shorter notation $SCC(e)$. Within this article, we use the RDFS entailment consequence relation as defined in its specification\footnote{\url{http://www.w3.org/TR/rdf-mt/}}.
%\end{req}

%\begin{figure}[h]
%\centering
%
%\begin{scriptsize}
%\subfigure[\scriptsize{SubClasses of Event}]{\label{fig:subclasses}\includegraphics[width=0.60\textwidth]{SubClasses}}
%\subfigure[\scriptsize{Meet Class}]{\label{fig:meetpattern}\includegraphics[width=0.39\textwidth]{meetpattern}}
%
%\subfigure[\scriptsize{Communication Class}]{\label{fig:communicationpattern}\includegraphics[width=0.40\textwidth]{communicationpattern}}
%\subfigure[\scriptsize{Murder Class}]{\label{fig:communicationpattern}\includegraphics[width=0.59\textwidth]{murderpattern}}
%
%\end{scriptsize}
% 
%\caption{Subclasses of the Generic Event. }
%\label{fig:eventclasses}
%\end{figure}

\subsection{Using Existing Data Models}
\label{sec:existing-subsection}
% ------Ankita -------------
In this part, we review a number of state-of-the-art event ontologies.

\paragraph{An ontology for \underline{L}inking \underline{O}pen \underline{D}escriptions of \underline{E}vents (LODE)}
In 2009 UC Berkeley introduced the LODE\footnote{\url{http://linkedevents.org/ontology/}} ontology. In this ontology, an event is defined as an action which takes place at a certain time at a specific location. It can be a historical action as well as a scheduled action. There were previous models \cite{Arndt2007}\cite{doerr2003cidoc} for representing historic events and scheduled events. Some of them represent both types of events (i.e., historical and scheduled), e.g., EventsML-G2\footnote{\url{http://www.itpc.org/EventsML}}.
The LODE ontology proposed to build an interlingua model, i.e., a model which encapsulates the overlap among different ontologies e.g., CIDOC CRM\footnote{\url{http://www.cidoc-crm.org/OWL/cidoc_v4.2.owl}}, ABC Ontology\footnote{\url{http://metadata.net/harmony/ABC/ABC.owl}}, Event Ontology\footnote{\url{http://motools.sourceforge.net/event/event.html}}, and EventsML-G2. This encapsulation is utilized to create a mapping among existing ontologies. LODE was introduced to publish historical events in a fine-grained manner as it assumes each event is a unique event even if it is a part of a series. Because the concept of sub-events does not exist in LODE, related events can be interlinked. This ontology helps us to link factual aspects of a historical event. A factual aspect is given by 'What happened' (event), 'Where did it happen' (atPlace), 'When did it happen' (atTime), 'Who was involved' (involvedAgent) \cite{shaw2009lode}.
%LODE
\begin{figure}[htb]
\centering
\includegraphics[width=0.8\columnwidth]{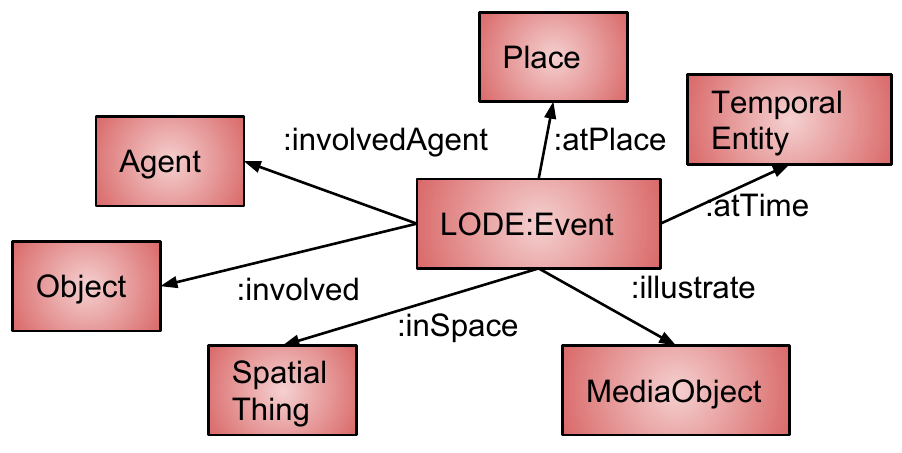} 
\caption{Schematic representation of LODE.}
\label{fig:exam}
\end{figure}
A visualization of LODE ontology is shown in Figure 2. We conclude that LODE meets (i) Requirement 1 as it defines a generic concept of the historic event, (ii) loosely Requirement 3 as it contains generic types for entities, e.g., Agent, SpatialThing, TemporalEntity, (iii) Requirement 4 as it includes necessary relations.
But LODE ontology fails to meet Requirement 2 as it does not include the publisher of the event (provenance).
Figure 3 depicts our running example in LODE.
\begin{figure}[htb]
\centering
\includegraphics[width=0.6\columnwidth]{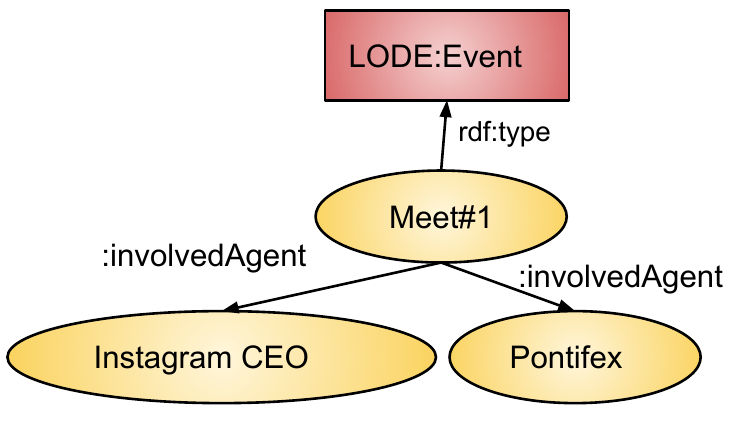} 
\caption{Representing the running example using LODE ontology.}
\label{fig:exam}
\end{figure}

\paragraph{\underline{S}imple \underline{E}vent \underline{M}odel (SEM)}
In 2011, SEM ontology was introduced from Vrije University and Delft. This ontology describes events as the central element in representing historical data\footnote{AGORA project \url{http://www.cs.vu.nl/˜schreiber/ projects/agora.html}
}, cultural heritage \cite{crofts2008definition}\cite{ruotsalo2007event} and multimedia \cite{westermann2007toward}. SEM is combined with a Prolog API  to create event instances without the background knowledge. This API also helps in connecting the created event instances to Linked Open Data. 
%SEM ontology mainly focuses on events occurring on the Web. 
SEM proposes a method to attain interoperability among datasets from different domains.
SEM strives to remove constraints to make it reusable by supporting weak semantics. 
%Most of the models present classify the same thing into different domains to cope up with the disarrangement of the Semantic Web as this allows reuse of the data present on Web. 
%In SEM the model is less constraining to make it more reusable which also helps for the applications which were not present beforehand. 
Thus, in SEM, the concept of event is specified as everything that happens \cite{van2011design}. 
%In SEM ontology, the individuals and types both are considered as classes. This helps to link it to various vocabularies. Also, the use of only rdfs:domain and rdfs:range to classes help in linking with classes available in other vocabularies without inheriting their constraints.
%SEM is classified in three groups\cite{van2011design}: 
%\begin{itemize}
%    \item Core Classes
%    \item Types
%    \item Constraints
%\end{itemize}
%In SEM, the Core Classes are of four types\cite{van2011design}:
%\begin{itemize}
%    \item sem:Event (what happened)
%    \item sem:Actor (who participated)
%    \item sem:Place (where the event has happened)
%    \item sem:Time (when the event has happened)
%\end{itemize}
%To specify the type of the core individuals, each core class is associated with sem:Type class.
%There are three types of sem:Constraints\cite{van2011design}: \begin{itemize}
%    \item sem:Role
%    \item sem:Temporary
%    \item sem:View
%\end{itemize}
%In SEM, properties are divided as:
%\begin{itemize}
%    \item sem:eventProperty
%    \item sem:type
%    \item sem:accordingTo's subproperties
%    \item sem:hasTimeStamp's subproperties
%\end{itemize}
%SEM from google drive
\begin{figure}[htb]
\centering
\includegraphics[width=0.8\columnwidth]{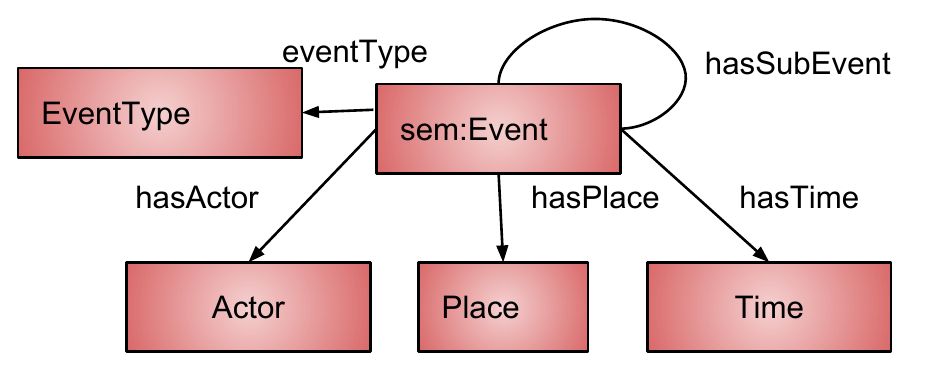} 
\caption{Schematic representation of SEM.}
\label{fig:sem}
\end{figure}
A schematic representation of SEM model is shown in \autoref{fig:sem} (summarized version). We conclude that SEM meets (i) Requirement 1 as it defines generic event, (ii) Requirement 3 as it specifies a type for entities, e.g., Actor, and (iii) Requirement 4 as it includes required properties. Similar to LODE ontology, SEM model fails to meet Requirement 2 as it does not include the publisher of events (provenance).
Fig 
%SEM-example from google drive
%\begin{figure}[htb]
%\centering
%\includegraphics[width=\columnwidth]{example1} 
%\caption{Representing the running example using SEM.}
%\label{fig:exam}
%\end{figure}
%Fig.5 depicts a running example of SEM model. This model has 2 Actors (i.e. Instagram CEO and Pontifex), one Event (i.e. meet).

\paragraph{dbpedia.org/ontology/Event} The DBpedia ontology\footnote{\url{http://wiki.dbpedia.org/services-resources/ontology}} defines the generic concept of event with a hierarchy which is broader, including lifecycle events (e.g. birth, death), natural events (e.g. earthquake, stormsurge), and  societal events (e.g. concert, election).
%The DBPedia is based on the data corpus from Wikipedia. The DBPedia helps in extracting structured information from Wikipedia using Semantic Web techniques. This also helps in linking Wikipedia data to other datasets present on the Web. The advantage of DBPedia is that the third party applications can import DBPedia datasets. As of now, DBpedia dataset contains information about more than 1.95M ”things”: 80K persons, 70k places, 35k music albums,  and 2k films \cite{auer2007dbpedia}. It also provides links to 657K images, 1.6M  external web pages, 180K RDF datasets, 207K Wikipedia categories and 75K YAGO categories. \cite{suchanek2007yago}. 
We conclude that DBpedia meets (i) Requirement 1 as it defines generic event, (ii) Requirement 3 as it specifies a type for entities, and (iii) Requirement 4 as it includes required properties. All these can be imported from other datasets present on the Web as DBpedia links to other datasets in an easy manner. Similar to LODE ontology and SEM model, DBpedia fails to meet Requirement 2 as it does not include the publisher of events (provenance).

\begin{figure}[htb]
\centering
\includegraphics[width=0.5\columnwidth]{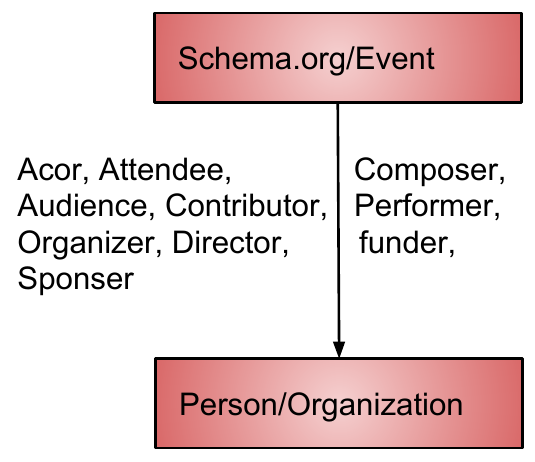} 
\caption{Schematic representation of Schema.org.}
\label{fig:schema.org}
\end{figure}

\paragraph{schema.org/Event}
Schema.org\footnote{\url{http://schema.org}}, a product of collaborative efforts by major companies (i.e., Google, Bing, Yahoo and Yandex)  \footnote{https://en.wikipedia.org/wiki/Schema.org},
presents similar generic concept of event\footnote{\url{http://schema.org/Event}}. It considers temporal as well as location aspects and additionally provides a limited hierarchy.
This hierarchy introduces types of events such as business events, sale events, and social events. The schemas in schema.org are set of these types which are associted with a set of properties. 
Furthermore, it considers multiple labels between the associated entity and the concept of the event (represented in \autoref{fig:schema.org}) such as actor and  contributor, which distinguishes the role of the associated entity.
Schema.org introduces hundreds of schemas for categories like movies, music, organizations, TV shows, products, places etc \footnote{https://googleblog.blogspot.com/2011/06/introducing-schemaorg-search-engines.html}. For Schema.org, an event is an instance taking place at a certain time and at a certain location. Like LODE, the repeated events are classified different events and thus keeping all the events unique even if it is a sub event. A schematic representation of Schema.org (summarized version) is shown in \autoref{fig:schema.org}.
%A running example of Schema.org is shown in Fig. 6
% schema-example from google drive

%\begin{figure}[htb]
%\centering
%\includegraphics[width=0.5\columnwidth]{schema.pdf} 
%\caption{Schematic representation of Schema.org.}
%\label{fig:schema.org}
%\end{figure}

We conclude that Schema.org meets (i) Requirement 1 as it defines generic event, (ii) Requirement 3 as it specifies a type for entities e.g Actor (as type Person), Location (as type Place), Organizer (as type Person), StartDate (as type Date or DateTime) etc.. and (iii) Requirement 4 as it includes required properties for every entities defined above. Like LODE, SEM and DBPedia, Schema.org also fails in meeting Requirement 2 as it can define or import publisher of the event (provenance). 

\paragraph{\underline{C}omprehensive \underline{Ev}ent \underline{O}ntology (CEVO)}
The CEVO ontology\footnote{\url{http://eventontology.org/}} relies on an abstract conceptualization of English verbs provided by Beth Levin \cite{levin_english_1993}.
Levin categorizes English verbs according to shared meaning and behavior.
CEVO ontology, which is a machine-readable format (i.e., RDF format) of Levin 's categorization, presents more than 230 event classes for over 3,000 English verbs individuals.
It organizes classes into semantically coherent event classes and event hierarchy, and notably, has an inventory of the corresponding lexical items. 
For example, \autoref{tab:threeVerbClasses} in the first column presents three event classes as (i) {\it Communication event\/} that corresponds to the event which causes transferring a message, (ii) {\it Meet event\/} which is an event related to group activities, and (iii) {\it Murder event\/} which is referring to an event that describing killing.
The second column of \autoref{tab:threeVerbClasses} represents the lexical items (i.e., verbs) having shared meaning and are under the umbrella of a common event. In other words, an appearance of one of these verbs shows the occurrence of its associated event.
For example, w.r.t. the running example, the appearance of the verb meet in the given tweet shows the occurrence of an event with the specific type `meet'.

\begin{table}[hpt]
	\centering
\begin{scriptsize}
\begin{tabular}{ l|l}
\hline
\textbf{Event Classes} 				&  \textbf{Associated Verbs} 		 	   \\ \hline

\multirow{4}{*} {\emph{Communication event}}	 	&     \emph{Say Verbs:} admit, allege, announce, articulate, assert,  \\ & communicate, confess, convey, declare. 	\\											&	mention, propose, recount, repeat, report, reveal, say, state.\\
										&	\emph{Tell Verbs:} ask, cite, pose, preach, quote, \\ & read, relay, show, teach, tell, write,demonstrate.	\\
										&	dictate, explain, explicate, narrate, teach. \\
     								 	 	 	  	 \hline

\multirow{1}{*} {\emph{Meet event}}	   &	battle, box, consult, debate, fight, meet, play, visit.	  \\  	\hline

\multirow{2}{*} {\emph{Murder event}}	 		&  assasinate, butcher, dispatch, eliminate, execute, slay. \\
										&  immolate, kill, liquidate, massacre, murder, slaughter.  		\\	
											 	 	  \hline 
						   					       
\end{tabular}
\end{scriptsize}

\caption{ Three samples of events from CEVO with their English verbs.}
\label{tab:threeVerbClasses}
\end{table}

The CEVO ontology can be employed for recognizing events and more interesting classifying them w.r.t. their specific type. Specifically, it unifies apparently disparate lexical items under a single event class. More importantly, this can prove 
critical in reducing the number of apparent features for classifiers and in the support of inference necessary for query response.

\subsection{Developing a Data Model}
\begin{figure*}[hbt]
\centering
\begin{scriptsize}
\subfigure[\scriptsize{SubClasses of Event}]{\label{fig:subclasses}\includegraphics[width=0.25\textwidth]{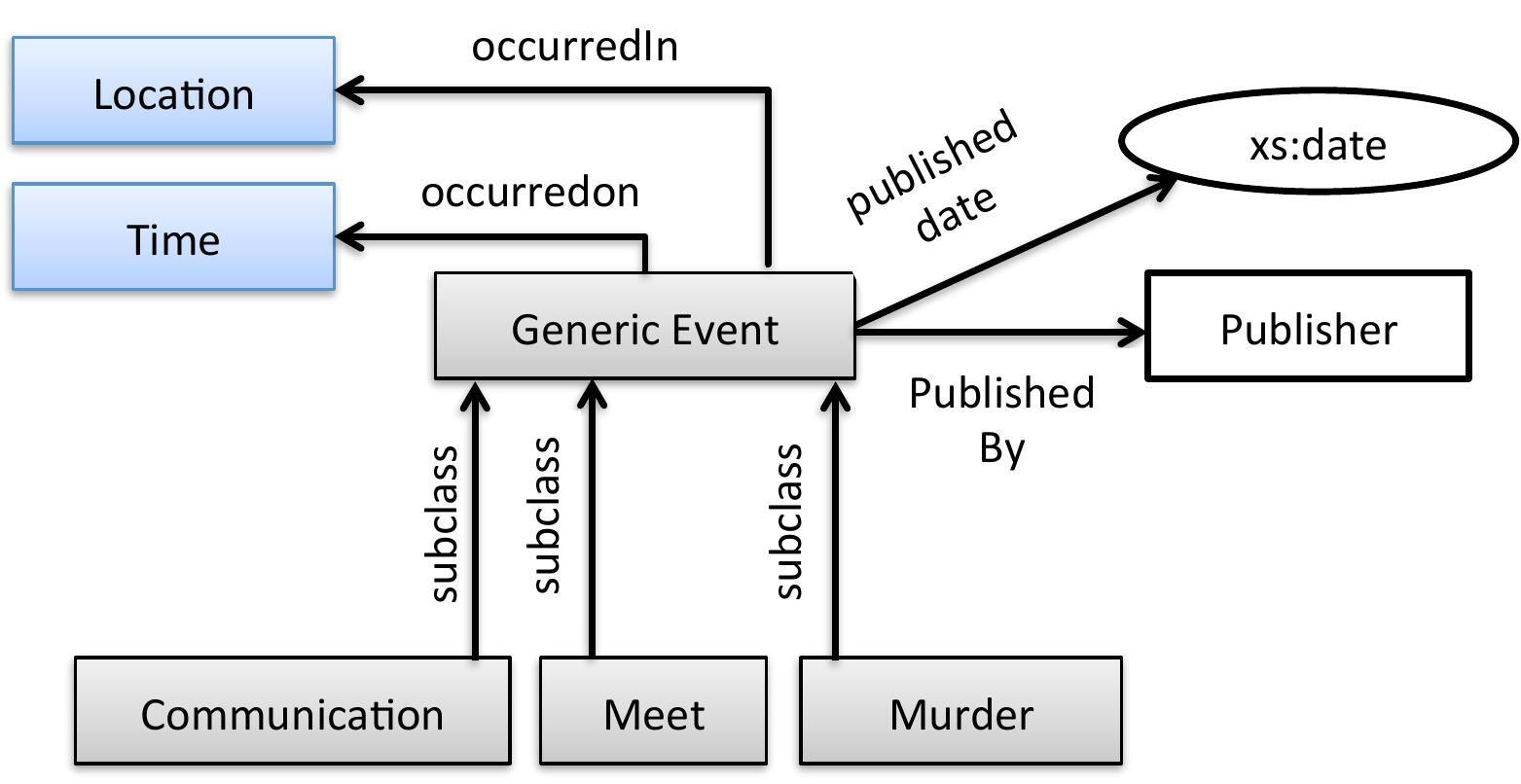}}
\subfigure[\scriptsize{Meet Class}]{\label{fig:meetpattern}\includegraphics[width=0.2\textwidth]{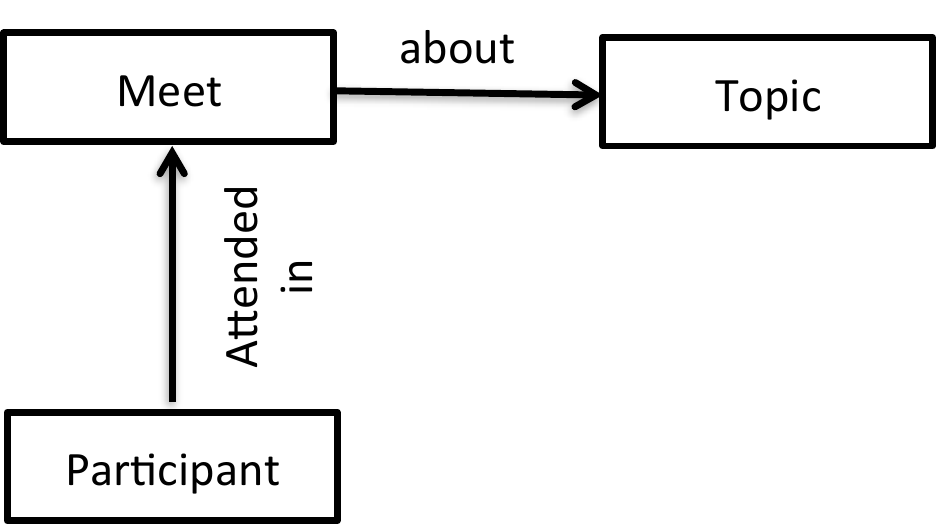}}
\subfigure[\scriptsize{Communication Class}]{\label{fig:communicationpattern}\includegraphics[width=0.25\textwidth]{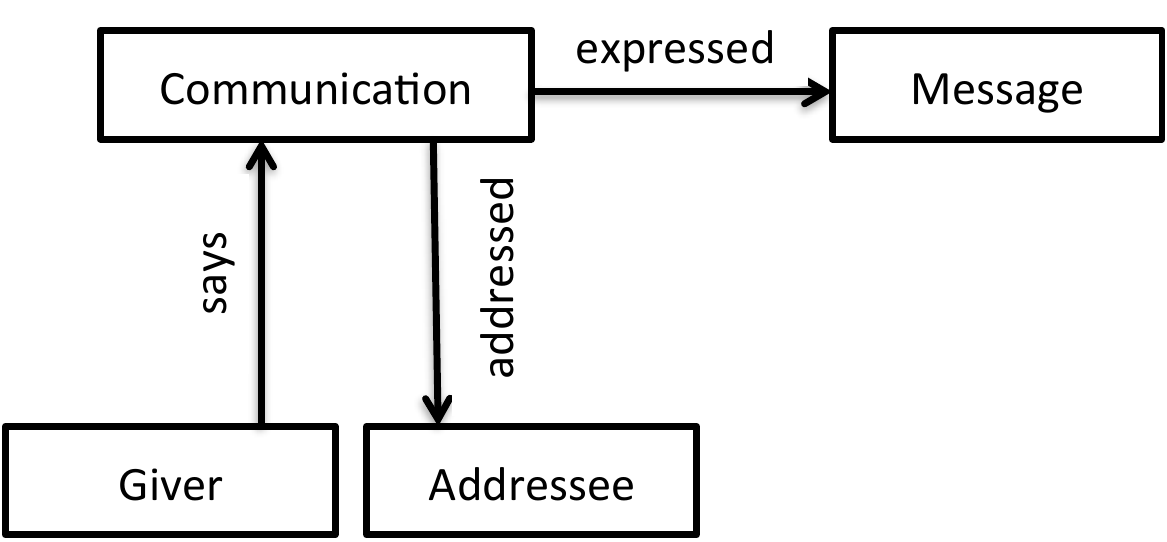}}
\subfigure[\scriptsize{Murder Class}]{\label{fig:communicationpattern}\includegraphics[width=0.25\textwidth]{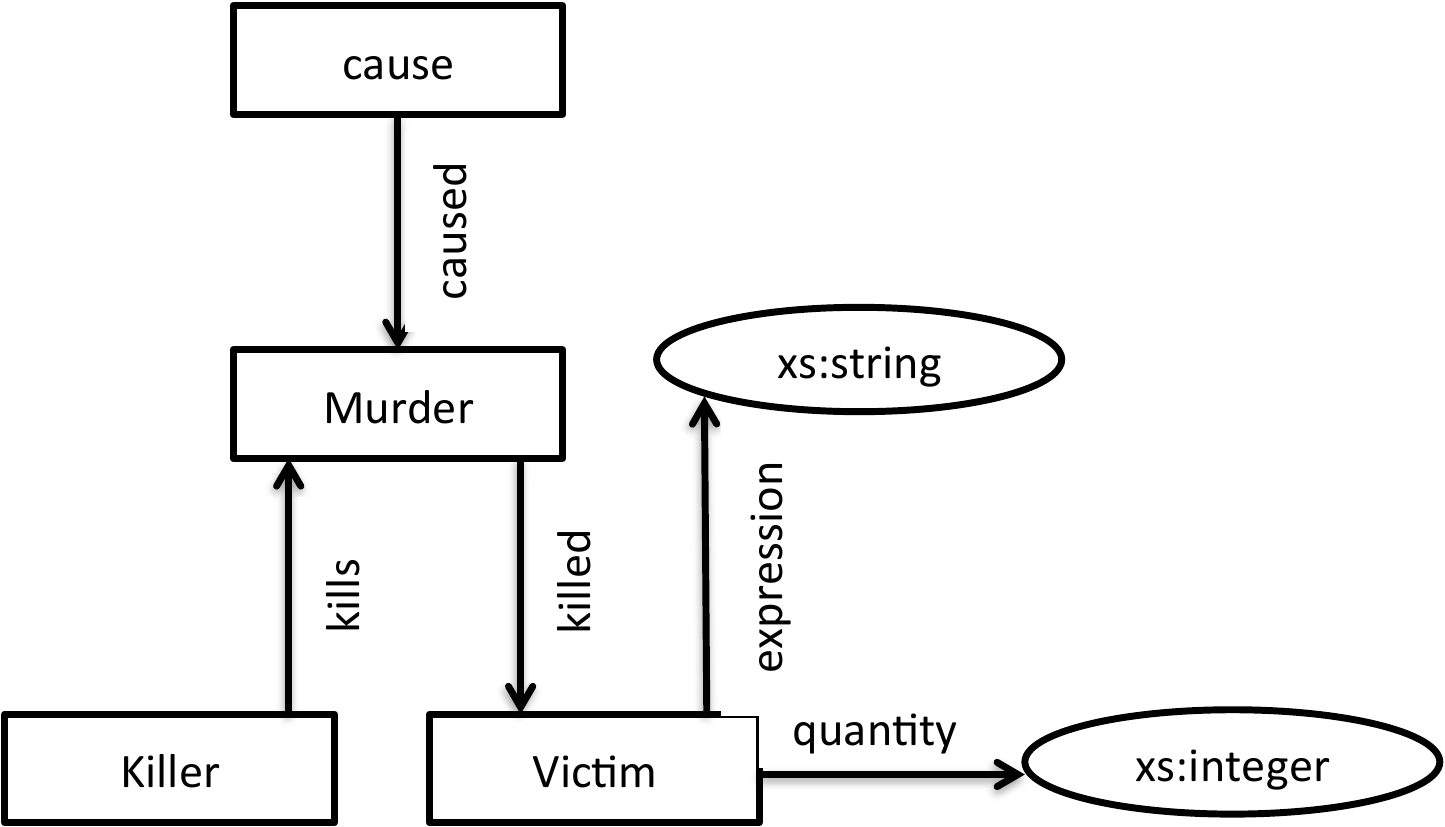}}
\end{scriptsize}
 
\caption{Schematic overview for generic event an three sub-events i.e. meet, communication, and murder. }
\label{fig:eventclasses}
\end{figure*}

The existing data models are basically coarse-grained. In case the domain or application requires a fine-grained data model, the existing data models can be extended.
For example, here we extended event data model from CEVO\footnote{\url{http://eventontology.org}} ontology for three specific events.
We take into account three subclasses (shown in Figure \ref{fig:subclasses}) as (i) class communication $C_{communication} \sqsubset GEvent$ that refers to any event transferring a message, (ii) class meet $C_{meet} \sqsubset GEvent$ that ranges over all group activities, and finally, (iii) class murder $C_{murder} \sqsubset GEvent$ 
that includes any reports of killing.

Furthermore, as Figure \ref{fig:subclasses} shows, the provenance information (e.g., publisher or date) is represented within the data model (default arguments for all events), to meet Requirement \autoref{req:prov}.
Figure \ref{fig:eventclasses}(b-d) represents parts of data model for sub-event classes (i.e., $C_{meet}, C_{communication}, C_{murder}$) in detail.
The type of all possible associated entities as well as their necessary relationships are represented within the data model. 
This meets the Requirements  \ref{req:type} and \ref{req:relation}.
For example, the \texttt{meet} event is associated with entities with type of \texttt{Participant} and \texttt{Topic} (i.e., topic discussed in the meeting).
Considering the sample of tweets in Table \ref{tab:tweetsamples}, the tweets no.1, no.4, and no.7 are instances of the event \texttt{Communication} with the mentions \texttt{tell, say, announce}.
The tweets no.2, no.5, no.8 are instances of the event \texttt{Meet} with the mentions \texttt{meet, visit}.
The tweets no3, no6, no9 are instances of the event \texttt{Murder} with the mention \texttt{kill}.
\autoref{fig:exam} demonstrates the running example within the developed data model.
This event has two participants (i.e. instagram CEO and Pontifex) along with a specific topic.

%#################################

\begin{figure}[htb]
\centering
\includegraphics[width=0.8\columnwidth]{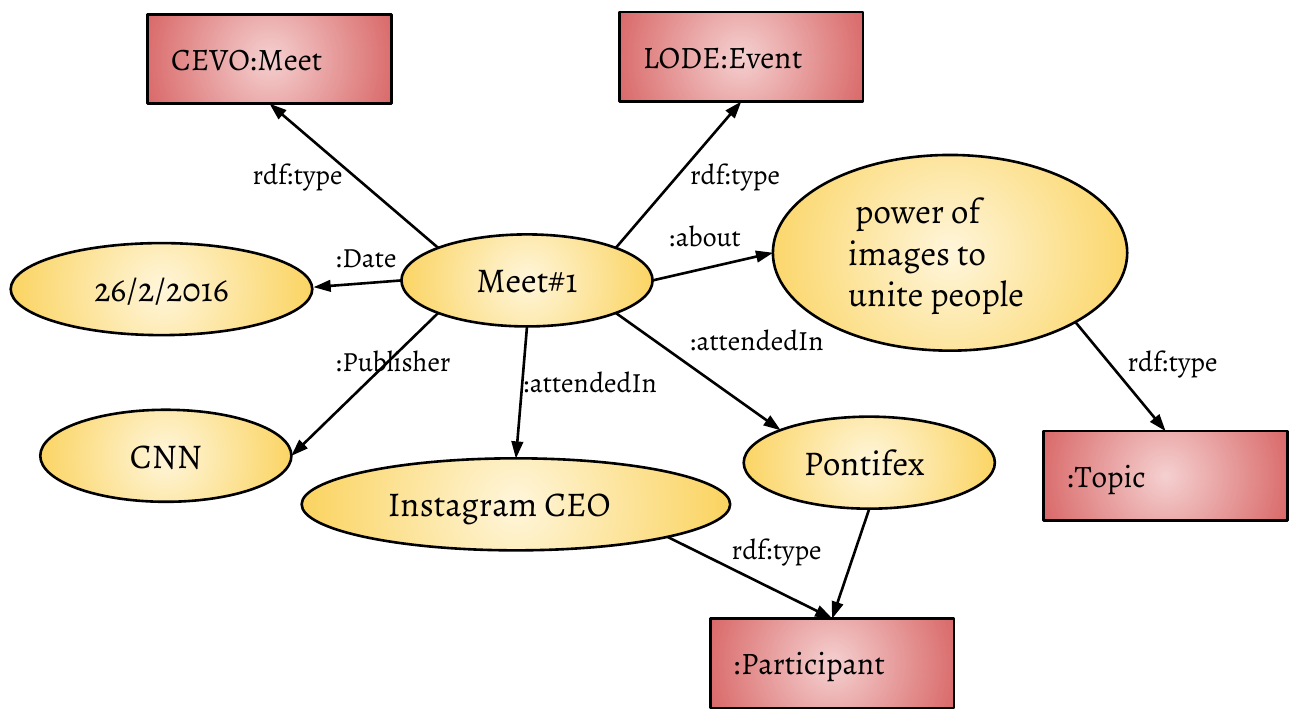} 
\caption{Representation of the running example using the model of \autoref{fig:eventclasses}.}
\label{fig:exam}
\end{figure}

%####################################

%---------------------------------------------------------------
\subsection{Using Singleton Property}
We can adopting the concept of a \emph{singleton property} introduced in~\cite{Nguyen:www2014} for modeling n-ary relations in the background data model.
Singleton properties replace  RDF reifications and enable efficient represention of statements about statements. 
Since news headlines contain both provenance information and multiple associated entities, SP is a suitable choice and furthermore, it enable systematic encoding of n-ary relations in terms of binary relations.

\begin{example}[Input/Output]
\begin{small}
Considering our running example which is about the occurrence of a {\tt meet} event with two participant entities \texttt{Instagram CEO} and \texttt{Pontifex} and the topic $t_1$.  The generated triples using singleton property are as follows:
\begin{lstlisting}
1. :Meet#1      singletonPropertyOf  	  :Meet.
2. :Instagram_CEO   :Meet#1                 :Pontifex.
3. :Meet#1              :about              :t1.
4. :Meet#1              :hasSource          :CNN.
5. :Meet#1              :extractedOn        `26/2/2106'.
6. :t1                   a	            :Topic. 
7. :t1                  :body	            `to discuss the power of images to unite people'.
\end{lstlisting}
\end{small}
%\end{verbatim}
\end{example}

\subsection{Event Annotation}
Events can be represented at different levels of granularity. 
The event annotation task potentially comprises of two subsequent tasks as follows: 
\begin{enumerate}[I]
    \item \emph{Event recognition:} Typically, event recognition  utilizes phrases and their parts of speech. Although, verbs are more common for distinguishing an event (e.g., \textit{`Obama met Merkel in Berlin'}), the other POS might reveal an event (e.g., \textit{`G8 meeting in Berlin'}).
    Furthermore, event recognition task  ecan beither open domain or closed domain. In the former one, collecting a lexicon of event phrases is more challenging rather than for the latter one.
    In any case, a learning approach (either supervised or semi-supervised) can be applied for determining whether or not a piece of text contains an event phrase or not. 
    \item \emph{Event classification:} This task is necessary in case the employed background data model considers the specific type of events as part of event annotation. In this case, event phrases have to be labeled by specific types of events using multi-class classifier trained for distinguishing the specific type of a given event. For example, the tweets no.2, no.5, no.8 of \autoref{tab:tweetsamples} have the specific type \textit{``meet''}.
\end{enumerate}

\subsection{Entity Annotation}
Entity annotation is a significant task for creating a knowledge graph of events. 
It can be challenging when we have a fine-grained background data model, which makes the task of semantic role labeling of entities necessary.  
Overall, the required tasks for fulfilling entity annotation are as follows:

\begin{enumerate}[I]
    \item \emph{Entity recognition:} This task specifies a chunk of text as an individual entity which plays a role in the occurred event. An entity mention can be explicit or implicit. Regarding explicit entities,  Named Entity Recognition (NER) tools can be used for open domain scenarios whereas  alternatives such as knowledge graphs, gazetteers, and domain dictionaries are 
    necessary for closed domain scenarios.
    E.g., for the tweet no.1 in \autoref{tab:tweetsamples}, the chunk \textit{`Michelle Obama'} is recognized as a named entity with the type person.
    \item \emph{Entity linking:} Entity linking can be attributed into two tasks, the first one \cite{DBpedia-spotlight}, which is required in our case, is about associating entity mentions in a given text to their appropriate corresponding entities in a given knowledge graph. Thus, it removes ambiguity. A textual mention of an entity might have a matching entity in the knowledge graph or not. In the former case, entity linking task is reduced to hook a suitable entity whereas in the latter case, it is required that a new IRI (i.e., International Resource Identifier) be minted and typed and then linked to the textual mention of the given entity. E.g., in the tweet no.1 of \autoref{tab:tweetsamples}, the named entity \textit{`Michelle Obama'} should be linked to the entity \texttt{dbr:Michelle\_Obama}\footnote{dbr is the prefix standing for \url{http://dbpedia.org/resource/}.}, when DBpedia is employed as the background knowledge graph.
    The second type of entity linking is about linking entities across knowledge graphs using \texttt{owl:sameAs} links. While the first task is required in the pipeline of developing an event knowledge graph, the second one is optional but can enhance quality and visibility of the underlying knowledge graph.

    \item \emph{Semantic role labeling:}
    Most of the existing event ontologies consider generic roles such as actor or agent for involved entities. 
    For fine-grained background data model, the semantic role labeling can be done.
    E.g., w.r.t. the tweet no.1 in \autoref{tab:tweetsamples}, the entity \textit{`Michelle Obama'} can be labelled by the generic role \emph{actor} employing LODE ontology or the specific role \emph{giver} applying the data model illustrated in \autoref{fig:communicationpattern}.   
    \item \emph{Entity disambiguation:} An entity mention in a text might be polysemous, thus linking  to the correct entity in the underlying knowledge graph requires a disambiguation phase.
    Furthermore, a single entity in multiple knowledge graphs might have various representations. Thus, interlinking them is challenging and requires a disambiguation phase as well \cite{NEDisambiguation2007,NEDisambiguation2010,usbeck2014agdistis}.
    E.g., w.r.t. the tweet no.7 in \autoref{tab:tweetsamples}, the named entity `Obama' is ambiguous as of whether it refers to `Michelle Obama' or `Barack Obama'. 
    Regarding context (i.e., the remaining part of the tweet), it likely refers to `Barack Obama'. 
 
    \item \emph{Implicit entity linking:} As we mentioned before, not all of the mentions of entities are explicit. For example, w.r.t. the running example, the chunk \emph{`Instagram CEO'} refers to the implicit entity `Kevin Systrom' who is the CEO of Instagram. The experiment performed in \cite{ESWC/ImplicitEntities} shows that 21\% entity mentions in movie domain and 40\% of entity mentions in Book domain are implicit. Inferring implicit entities depends on capturing context as well as respecting time intervals.
\end{enumerate}

\subsection{Interlinking Events}
The tasks described above have been considered independently before.  
The interlinking requirement, which has not hyet been adequately explored, comes from the two inherent facts of events as follows:
\begin{enumerate}[I]
\item A single event might be reported by various publisher sources using different expressions. Thus, it is necessary to identify same events across various publisher sources, and then interlink them using \texttt{owl:sameAs} or \texttt{skos:related} links. 
\item Events have an evolutionary nature in the sense that more information is added with time.
Thus, it is essential to spot an event and its subsequent events reported to either complement the original event or reflect its causes or consequences.
To interlink such events, \texttt{skos:related} can be utilized.
\end{enumerate}

\paragraph{Publishing Event Triples According to Linked Data Principles} 
The recognized events, entities and relations have to be published according to  principles of LOD, RDF and the employed background data model.
To maintain the knowledge graph's consistency and coherence, the generated triples must be de-duplicated, validated and  assigned URIs disambiguated.
The minted URI should be dereferenceable and interlinked to external RDF data sources. 
%We address this in a manuscript in preparation.  
%WOW.  I THINK THIS WILL BE HARDER THAN YOU THINK.  %WE WILL HAVE TO DETERMINE WHEN THE SAME HEADLINE REFERS TO THE SAME OR DIFFERENT EVENTS. VLS

\section{Related Work}
\label{sec:relatedwork}
%\todoiteminlinedone{VLS}{Saeedeh}{I'm still not getting the point.  Do you want to say the existing work is flawed?  Domain specific?  Or perhaps that you build on it? We need to know the point!}
%\todoiteminline{VLS}{Saeedeh}{The purpose of this section is not clear.  You provide an inventory of methods but without comparison to the contribution of your approach.}
%\todoiteminlinedone{Amit}{Saeedeh}{I think talking about a holistic framework is asking for a trouble-- never overpromise and use broad terms like "holistic" -- you have not discussed what is holistic! Instead keep to the targeted message-- one that is more focused on what is evaluated and demoed. This is the kind of thing we could write much better if there is time.}

Overall, there is a lack of a holistic view on event extraction from free text and subsequently developing a knowledge graph from it.
In this paper, we presented the full pipeline containing the required tasks such as (i) agreeing upon a data model, (ii) event annotation, (iii) entity annotation
and (iv) interlinking events.
The majority of previous research is either domain-specific or event-specific and
do not undertake the full pipeline (e.g., limited to only event and entity extraction).
We  have provided a visionary review of the full pipeline which is merely applicable to any domain.
%While the proposed framework is simple applicable to any particular event or domain.
In the following, we initially refer to research approaches for n-ary relation extraction on particular domains, then we refer the prominent approaches of binary relation extraction. 
We end by citing successful attempts at triple extraction from structured and semi-structured data sources. 

The work presented in \cite{McDonald2005} introduces complex relations as n-ary relations between n-typed entities. 
It proposes to factorize all complex relations into a set of binary relations. Then, a classifier is trained to recognize related entities of binary relations.
After identifying all pairs of related entities for binary relations, it reconstructs the complex relation using a simple graph creation approach.
Another domain for extracting n-ary relations is protein-protein interactions in the biomedical literature \cite{miyao2009evaluating,ono2001automated,xiao2005protein}.
They first identify protein mentions in text and then recognize interaction relations before finally extracting interactions. 
The approaches employed for protein-protein interactions can be divided into three groups: (i) graph-based approaches (e.g. co-occurrence graph),
(ii) rule-based approaches and (iii) learning approaches (e.g. maximum entropy).
%\todoiteminline{Saeede}{VLS}{ Saeede: I just mentioned the earlier work which have been done for extracting n-ary relations --------- Whats the point here? Domain specific techniques? If so, how does that fit with your effort?}

The other category of event extraction is based on binary relation extraction. NELL: Never-Ending Language Learning \cite{NEL2015} is a learning agent that extracts new facts using existing binary relations in its knowledge base. It was initiated in 2010 using a couple of seed binary relations but after years of running has become self-learning. A notable feature of NELL is its dynamic approach for extracting facts, as it refreshes beliefs in its knowledge base and removes the incorrect or old ones.
Linked Open Data as a valuable source of diverse ontologies also can be employed for extracting either new facts or new relations.
The framework proposed in \cite{rdflive,BOA2013} extracts facts using binary relations from DBpedia as background knowledge. 
In contrast to NELL, it initially identifies Named Entities and their type on plain text, then it tries to infer mentions of relation expression to properties in DBpedia (e.g. taking the domain and range of properties into account). % with confidence rate.
Open Information Extraction \cite{banko2008tradeoffs} is another extraction framework that is not limited to any predefined relation set. 
%\todoiteminline{Saeede}{TKP}{Saeede: I added some example for NELL and BOA framework in the challenge 1 section, do you think that shall I repeat here as well? OVERALL, CAN WE INCLUDE SOME EXAMPLE THAT 
%CONTRASTS OUR APPROACH WITH OTHERS TO BETTER ARGUE OUR CASE. AT THE MINIMUM CONTRASTING WITH REIFICATION OR ANY OTHER REFERENCED SCHEMES.}
%\todoiteminlinedone{VLS}{Saeedeh}{Whatever we do, the purpose of the review needs clarification.}
Furthermore, extracting triples from structured as well as semi-structured data sources has received adequate attention in the past, especially, DBpedia \cite{dbpedia_iswc} and LinkedGeo Data \cite{linkedgeodata} that leverage the loose structure of data for extraction.
Another example is the work  \cite{buhmann2014web} which presents a holistic approach for extraction of RDF from templated websites.

\section{Conclusion and Future Work}
\label{conclusion}

In this paper, we presented the initial version of our framework for the real-time extraction of events.
This framework is part of our project HeadEx for developing a knowledge graph of interlinked events.
We presented the requirements for choosing a data model representing events and their arguments. 
We reviewed the existing data models which have been employed by the state-of-the-art applications.
Furthermore, we outlined the required tasks for annotating events as well entities.
Then, the interlinking strategies were discussed.
As a proof-of-concept, we followed a case study of news headlines on Twitter.
For our future agenda, we plan to develop the envisioned pipeline containing all the required tasks by either implementing new components or integrating the existing ones.

%containing three distinct n-ary relations.
%The results of our experiments are promising and can be used to create timely and 
%structured news headlines dataset.
%Then, we use learning approaches, employing proposed syntactic features derived from  parsing,
%to extract information respecting the data model.
%We evaluated the event recognition and entity extraction tasks employing three different classifier algorithms along with various settings over the engaged features.
%Although we employed multilabel classifiers, the observed results are promising.
%Still there is an opportunity to apply sequenced-based classifiers such as conditional random field for further improvements.
%Improving learning approaches is one of our future aims.
%Furthermore, we currently focus on the challenges for deduplicating and validating triples and then assigning disambiguated URIs in order to publish this knowledge base on Linked Data. 

%\textbf{Acknowledgments} 
%We acknowledge partial support from the National Science
%Foundation (NSF) award: EAR 1520870: Hazards SEES: Social and Physical
%Sensing Enabled Decision Support for Disaster Management and Response. Any
%opinions, findings, and conclusions/recommendations expressed in this material
%are those of the author(s) and do not necessarily reflect the views of the NSF.

\bibliographystyle{plain}
\bibliography{bib/newsHeadline}
\end{document}